\documentclass[letterpaper]{article} 
\usepackage{aaai23}  
\usepackage{times}  
\usepackage{helvet}  
\usepackage{courier}  
\usepackage[hyphens]{url}  
\usepackage{graphicx} 
\usepackage{natbib}  
\usepackage{caption} 
\usepackage{algorithm}
\usepackage{algorithmic}
\usepackage{textcomp}
\usepackage{gensymb}
\usepackage{amsmath}
\usepackage{amssymb}
\usepackage{booktabs}
\usepackage{threeparttable}
\usepackage[table,xcdraw]{xcolor}
\usepackage{comment}
\usepackage{arydshln}
\usepackage{makecell}
\usepackage{subcaption}
\usepackage[super]{nth}
\usepackage{multirow}
\usepackage{threeparttable}
\usepackage[super]{nth}

\let\svthefootnote\thefootnote
\newcommand\blankfootnote[1]{%
  \let\thefootnote\relax\footnotetext{#1}%
  \let\thefootnote\svthefootnote%
}

\usepackage{newfloat}
\usepackage{listings}
\DeclareCaptionStyle{ruled}{labelfont=normalfont,labelsep=colon,strut=off} 
\floatstyle{ruled}
\newfloat{listing}{tb}{lst}{}
\floatname{listing}{Listing}
\title{LidarMultiNet: Towards a Unified Multi-Task Network for LiDAR Perception}
\author {
    Dongqiangzi Ye\textsuperscript{\rm 1}\equalcontrib,
    Zixiang Zhou\textsuperscript{\rm 1,2}\equalcontrib\thanks{Work done during an internship at TuSimple.},
    Weijia Chen\textsuperscript{\rm 1}\equalcontrib,
    Yufei Xie\textsuperscript{\rm 1}\equalcontrib,\\
    Yu Wang\textsuperscript{\rm 1},
    Panqu Wang\textsuperscript{\rm 1},
    Hassan Foroosh\textsuperscript{\rm 2}
}
\affiliations {
    \textsuperscript{\rm 1} TuSimple\\
    \textsuperscript{\rm 2} University of Central Florida
}

\newcommand{\lidarmultinet}{\mbox{LidarMultiNet }}

\begin{document}

\maketitle

\begin{abstract}
LiDAR-based 3D object detection, semantic segmentation, and panoptic segmentation are usually implemented in specialized networks with distinctive architectures that are difficult to adapt to each other. This paper presents \textbf{LidarMultiNet}, a LiDAR-based multi-task network that unifies these three major LiDAR perception tasks. Among its many benefits, a multi-task network can reduce the overall cost by sharing weights and computation among multiple tasks. However, it typically underperforms compared to independently combined single-task models. The proposed LidarMultiNet aims to bridge the performance gap between the multi-task network and multiple single-task networks. At the core of LidarMultiNet is a strong 3D voxel-based encoder-decoder architecture with a Global Context Pooling (GCP) module extracting global contextual features from a LiDAR frame. Task-specific heads are added on top of the network to perform the three LiDAR perception tasks. More tasks can be implemented simply by adding new task-specific heads while introducing little additional cost. A second stage is also proposed to refine the first-stage segmentation and generate accurate panoptic segmentation results. LidarMultiNet is extensively tested on both Waymo Open Dataset and nuScenes dataset, demonstrating for the first time that major LiDAR perception tasks can be unified in a single strong network that is trained end-to-end and achieves state-of-the-art performance. Notably, LidarMultiNet reaches the official \textbf{\nth{1}} place in the Waymo Open Dataset 3D semantic segmentation challenge 2022 with the highest mIoU and the best accuracy for most of the 22 classes on the test set, using only LiDAR points as input. It also sets the new state-of-the-art for a single model on the Waymo 3D object detection benchmark and three nuScenes benchmarks.

\end{abstract}

\section{Introduction}

\begin{figure}[t]
    \centering
    \begin{subfigure}[b]{0.49\linewidth}
         \frame{\includegraphics[width=\textwidth]{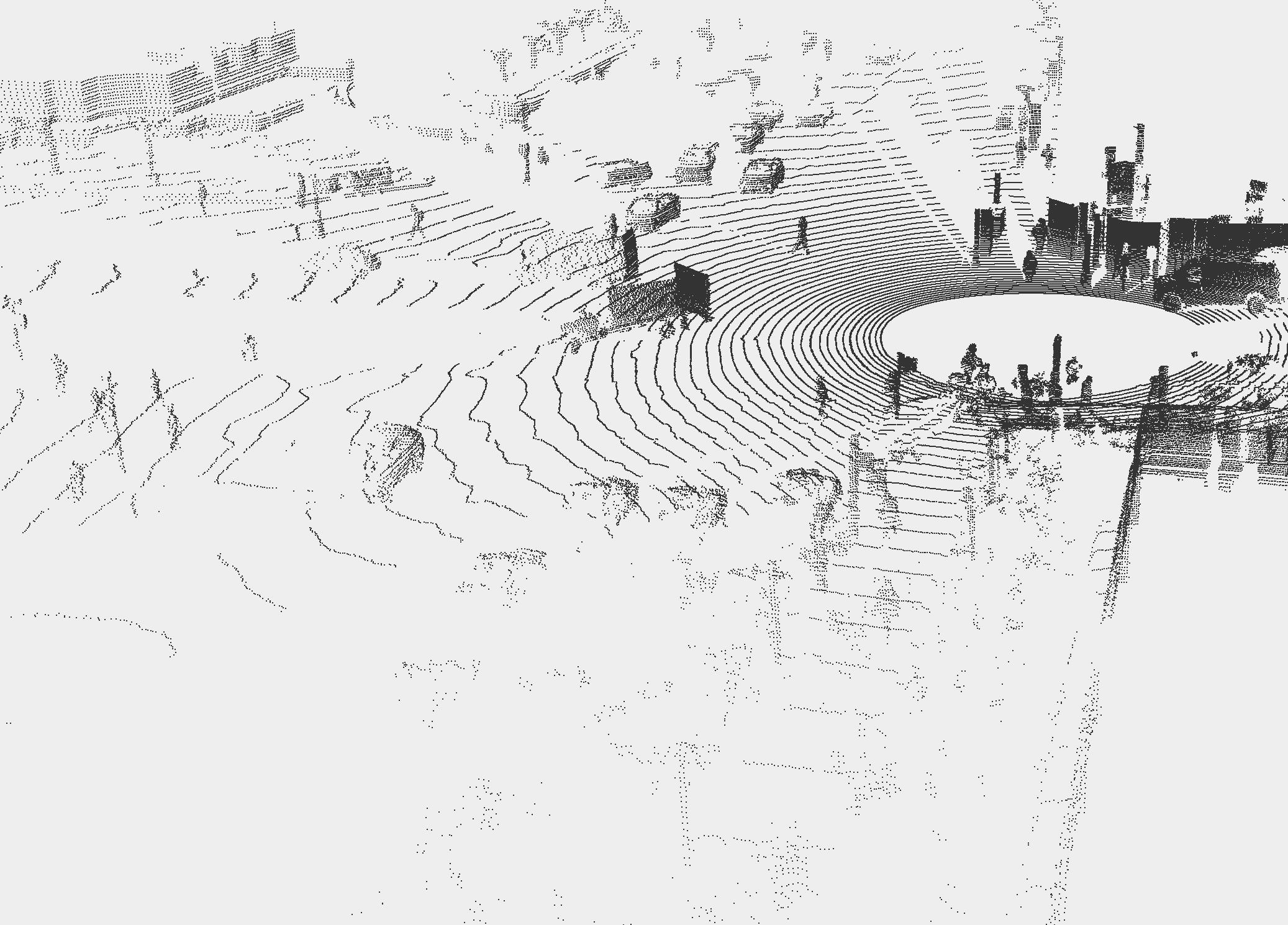}}
        \caption{}
    \end{subfigure}
    \begin{subfigure}[b]{0.49\linewidth}
        \frame{\includegraphics[width=\textwidth]{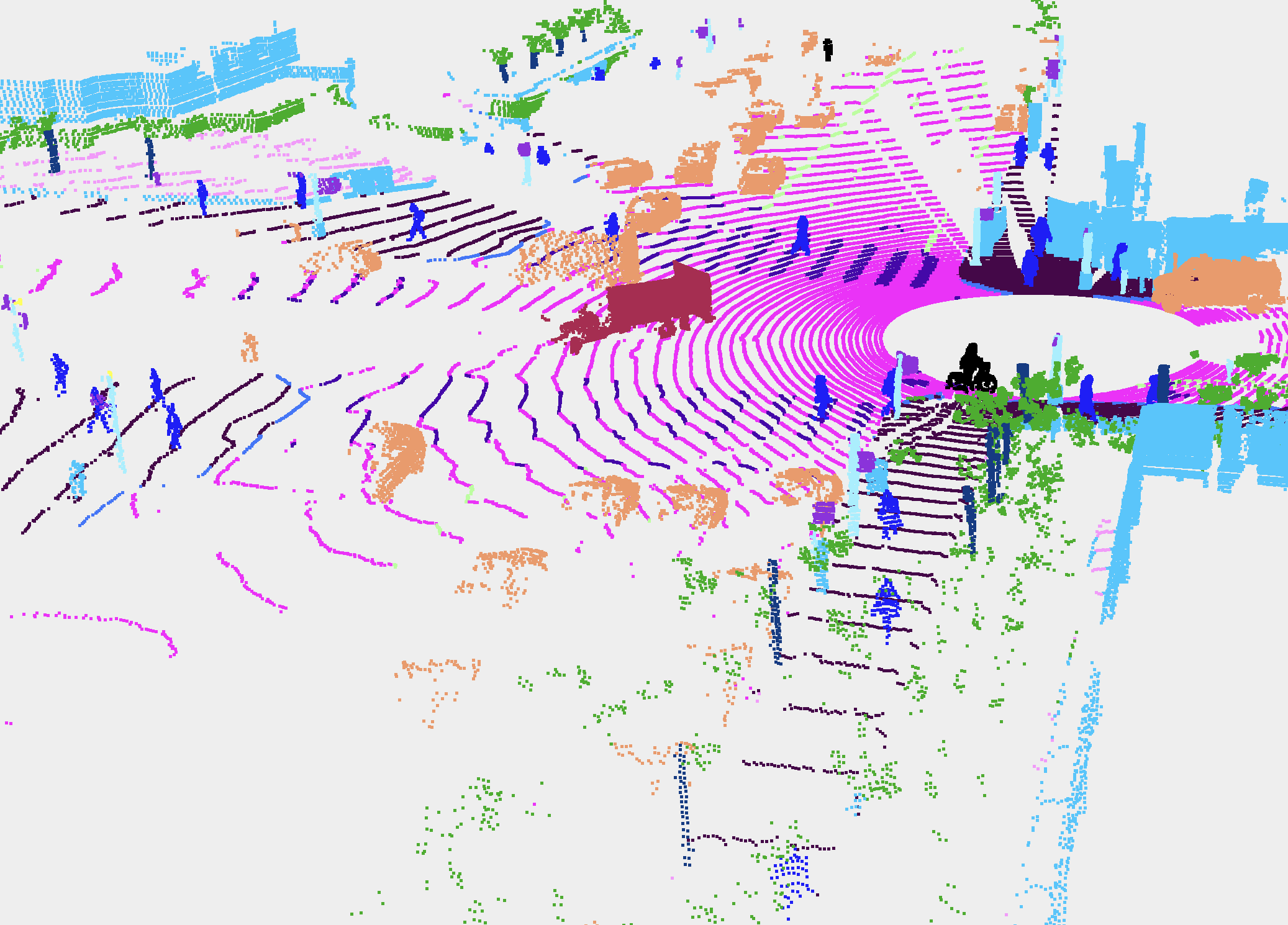}}
        \caption{}
    \end{subfigure}
    \begin{subfigure}[b]{0.49\linewidth}
        \frame{\includegraphics[width=\textwidth]{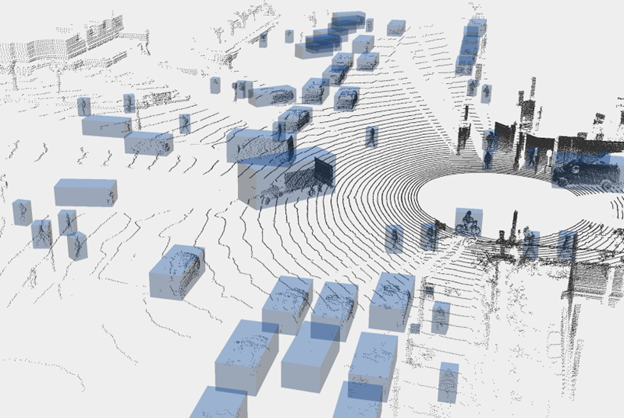}}
        \caption{}
    \end{subfigure}
    \begin{subfigure}[b]{0.49\linewidth}
        \frame{\includegraphics[width=\textwidth]{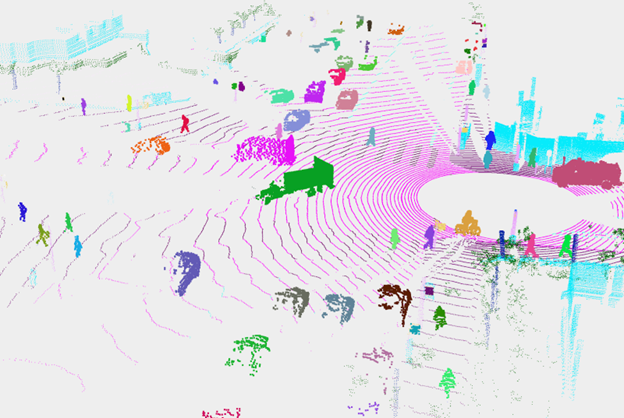}}
        \caption{}
    \end{subfigure}
    \caption{Our LidarMultiNet takes LiDAR point cloud (a) as input and performs simultaneous 3D semantic segmentation (b), 3D object detection (c), and panoptic segmentation (d) in a single unified network.}
    \label{fig:example}
\end{figure}

LiDAR plays a major role in the field of autonomous driving. With the release of several large-scale multi-sensor datasets, (\textit{e.g.} the Waymo Open Dataset~\cite{sun2020scalability} and the nuScenes ~\cite{caesar2020nuscenes}) datasets, collected in real self-driving scenarios, LiDAR-based perception algorithms have significantly advanced in recent years. Thanks to the advancement of sparse convolution~\cite{yan2018second,choy20194d}, voxel-based LiDAR perception methods~\cite{yin2021center} have become predominant on major 3D object detection and semantic segmentation benchmarks, and outperform their point-based, pillar-based, or projection-based counterparts~\cite{fan2021rangedet, lang2019pointpillars, qi2017pointnet} by a large margin in terms of both accuracy and efficiency. In voxel-based LiDAR perception networks, standard 3D sparse convolution is usually used in tandem with submanifold sparse convolution~\cite{submanifold2017}. Since standard 3D sparse convolution dilates the sparse features and increases the number of active sites, it is usually only applied as downsampling layer at each stage of the encoder followed by the submanifold sparse convolution layers. The submanifold sparse convolution maintains the number of active sites but it limits the information flow~\cite{Chen2022focalsparse} and the receptive field. However, a large receptive field is necessary to exploit the global contextual information, which is critical for 3D segmentation tasks.

In LiDAR-based perception, 3D object detection, semantic segmentation, and panoptic segmentation are usually implemented in distinct and specialized network architectures~\cite{yin2021center, Zhang_2020_CVPR, Zhou2021CVPR, zhu2021cylindrical, Cheng_2021_CVPR}, which are task-specific and difficult to adapt to other LiDAR perception tasks.  Multi-task networks~\cite{multinet2016, lidarmtl2021}, unify closely-related tasks by sharing the weights and computation among them, and therefore expected to improve the performance of individual tasks while reducing the overall computational cost. However, so far prior LiDAR multi-task networks have been underperforming compared to their single-task counterparts and have been failing to demonstrate state-of-the-art performance~\cite{lidarmtl2021}. As a result, single-task networks are still predominant in major LiDAR perception benchmarks. In this paper, we bridge the gap between the performance of single LiDAR multi-task networks and multiple independent task-specific networks. Specifically, we propose to unify 3D semantic segmentation, 3D object detection, and panoptic segmentation in a versatile network that exploits the synergy between these tasks and achieves state-of-the-art performance, as shown in Figure~\ref{fig:example}. 

Our main contributions are four-fold, and are summarized below:
\begin{itemize}
    \item We present a novel voxel-based LiDAR multi-task network that unifies three major LiDAR perception tasks and can be extended for new tasks with little increase in the computational cost by adding more task-specific heads.
    \item We propose a Global Context Pooling (GCP) module to improve the global feature learning in the encoder-decoder network based on 3D sparse convolution.
    \item We introduce a second-stage refinement module to refine the first-stage semantic segmentation of the foreground \textit{thing} classes and produce accurate panoptic segmentation results.
    \item We demonstrate start-of-the-art performance for \lidarmultinet on 5 major LiDAR benchmarks. Notably, \lidarmultinet reaches the official \nth{1} place in the Waymo 3D semantic segmentation challenge 2022. \lidarmultinet reaches the highest mAPH L2 for a single model on the Waymo 3D object detection benchmark. On the nuScenes semantic segmentation and panoptic segmentation benchmarks, \lidarmultinet outperforms the previously published state-of-the-art methods. On the nuScenes 3D object detection benchmark, LidarMultiNet sets a new standard for state-of-the-art performance in LiDAR-only non-ensemble methods. 
\end{itemize}

\section{Related Work}
\noindent\textbf{LiDAR Detection and Segmentation}
One key challenge for LiDAR perception is how to efficiently encode the large-scale sparsely distributed point cloud into a uniform feature representation. The common practice is transforming the point cloud into a discretized 3D or 2D map through a 3D voxelization~\cite{zhou2018voxelnet,zhu2021cylindrical}, Bird's Eye View (BEV) projection~\cite{yang2018pixor,lang2019pointpillars,Zhang_2020_CVPR}, or range-view projection~\cite{wu2018squeezeseg,sun2021rsn}. 
State-of-the-art LiDAR 3D object detectors~\cite{yin2021center} typically project the 3D sparse tensor into a dense 2D BEV feature map and perform the detection on the BEV space. In contrast, LiDAR segmentation requires predicting the point-wise labels, hence a larger features map is needed to minimize the discretization error when projecting the voxel labels back to the points. Many methods~\cite{tang2020searching,xu2021rpvnet,ye2021drinet} also combine the point-level features with voxel features to retain the fine-grained features in a multi-view fusion manner. 

In LiDAR-based 3D object detection, anchor-free detectors~\cite{yin2021center} are predominant on major detection benchmarks and widely adopted for their efficiency. Our LidarMultiNet adopts the anchor-free 3D detection heads, which are attached to its 2D branch. 

A second stage is often used in the detection framework~\cite{shi2020pv,yin2021center,li2021lidar,Sheng2021ICCV} to improve the detection accuracy through an RCNN-style network. It processes each object separately by extracting the features based on the initial bounding box prediction for refinement. LidarMultiNet adopts a second segmentation refinement stage based on the detection and segmentation results of the first stage.

\noindent\textbf{LiDAR Panoptic Segmentation}
Recent LiDAR panoptic segmentation methods~\cite{Zhou2021CVPR,hong2021lidar,razani2021gp} usually derive from the well-studied segmentation networks~\cite{Zhang_2020_CVPR,zhu2021cylindrical,Cheng_2021_CVPR} in a bottom-up fashion. This is largely due to the loss of height information in the detection networks, which makes them difficult to adjust the learned feature representation to the segmentation task. This results in two incompatible designs for the best segmentation~\cite{xu2021rpvnet} and detection~\cite{yin2021center} methods. According to ~\cite{fong2022panoptic}, end-to-end LiDAR panoptic segmentation methods still underperform compared to independently combined detection and segmentation models. In this work, our model can perform simultaneous 3D object detection and semantic segmentation and trains the tasks jointly in an end-to-end fashion.

\noindent\textbf{Multi-Task Network}
 Multi-task learning aims to unify multiple tasks into a single network and train them simultaneously in an end-to-end fashion. MultiNet~\cite{multinet2016} is a seminal work of image-based multi-task learning that unifies object detection and road understanding tasks in a single network. In LiDAR-based perception, LidarMTL~\cite{lidarmtl2021} proposed a simple and efficient multi-task network based on 3D sparse convolution and deconvolutions for joint object detection and road understanding. In this work, we unify the major LiDAR-based perception tasks in a single, versatile, and strong network.

\begin{figure*}
\centering
\includegraphics[width=1.0\textwidth]{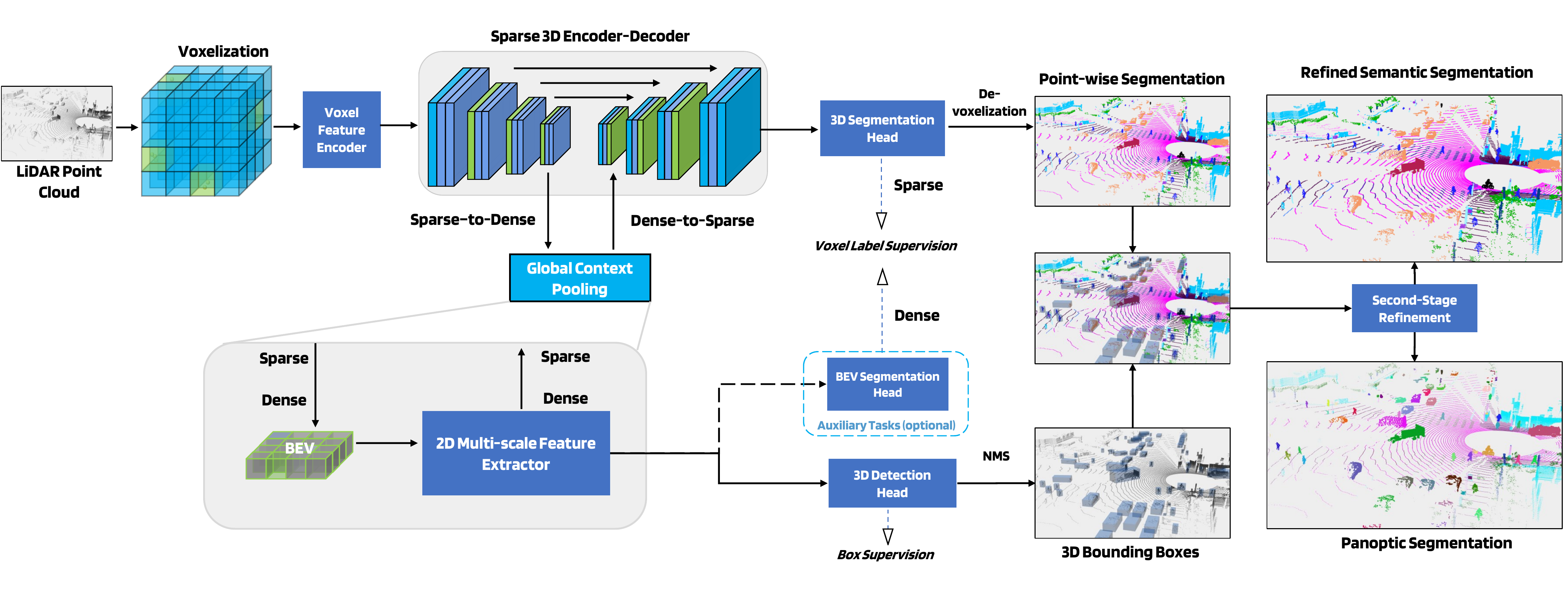}
\caption{\textbf{Main Architecture of the LidarMultiNet}. At the core of our network is a 3D encoder-decoder based on 3D sparse convolution and deconvolutions. In between the encoder and the decoder, a Global Context Pooling (GCP) module is applied to extract contextual information through the conversion between sparse and dense feature maps and via a 2D multi-scale feature extractor. The 3D segmentation head is attached to the decoder and its predicted voxel labels are projected back to the point level via a de-voxelization step. Meanwhile, the 3D detection head and auxiliary BEV segmentation head are attached to the 2D BEV branch. The 2nd-stage produces the refined semantic segmentation and the panoptic segmentation results.}
\label{fig:architecture}
\end{figure*}

\section{LidarMultiNet}

Given a set of LiDAR point cloud $P=\{p_{i}|p_{i}\in\mathbb{R}^{3+c}\}_{i=1}^{N}$, where $N$ is the number of points and each point has $3+c$ input features, the goals of the LiDAR object detection, semantic segmentation, and panoptic segmentation tasks are to predict the 3D bounding boxes, point-wise semantic labels $L_{sem}$ of $K$ classes, and panoptic labels $L_{pan}$, respectively. Compared to semantic segmentation, panoptic segmentation additionally requires the points in each instance to have a unique instance id.

\subsection{Main Architecture}
The main architecture of LidarMultiNet is illustrated in Figure~\ref{fig:architecture}. A voxelization step converts the original unordered LiDAR points to a regular voxel grid. A Voxel Feature Encoder (VFE) consisting of a Multi-Layer Perceptron (MLP) and max pooling layers is applied to generate enhanced sparse voxel features, which serve as the input to the 3D sparse U-Net architecture. Lateral skip-connected features from the encoder are concatenated with the corresponding voxel features in the decoder. A Global Context Pooling (GCP)~\cite{dongqiangzi2022lidarmultinet} module with a 2D multi-scale feature extractor bridges the last encoder stage and the first decoder stage. 3D segmentation head is attached to the decoder and outputs voxel-level predictions, which can be projected back to the point level through the de-voxelization step. Heads of BEV tasks, such as 3D object detection, are attached to the 2D BEV branch. Given the detection and segmentation results of the first stage, the second stage is applied to refine semantic segmentation and generate panoptic segmentation results.

The 3D encoder consists of 4 stages of 3D sparse convolutions with increasing channel width. Each stage starts with a sparse convolution layer followed by two submanifold sparse convolution blocks. The first sparse convolution layer has a stride of 2 except at the first stage, therefore the spatial resolution is downsampled by 8 times in the encoder. The 3D decoder also has 4 symmetrical stages of 3D sparse deconvolution blocks but with decreasing channel width except for the last stage. We use the same sparse convolution key indices between the encoder and decoder layers to keep the same sparsity of the 3D voxel feature map.

For the 3D object detection task, we adopt the detection head of the anchor-free 3D detector CenterPoint~\cite{yin2021center} and attach it to the 2D multi-scale feature extractor. Besides the detection head, an additional BEV segmentation head also can be attached to the 2D branch of the network, providing coarse segmentation results and serving as an auxiliary loss during the training. 

\begin{figure}
    \centering
    \includegraphics[width=1.0\linewidth]{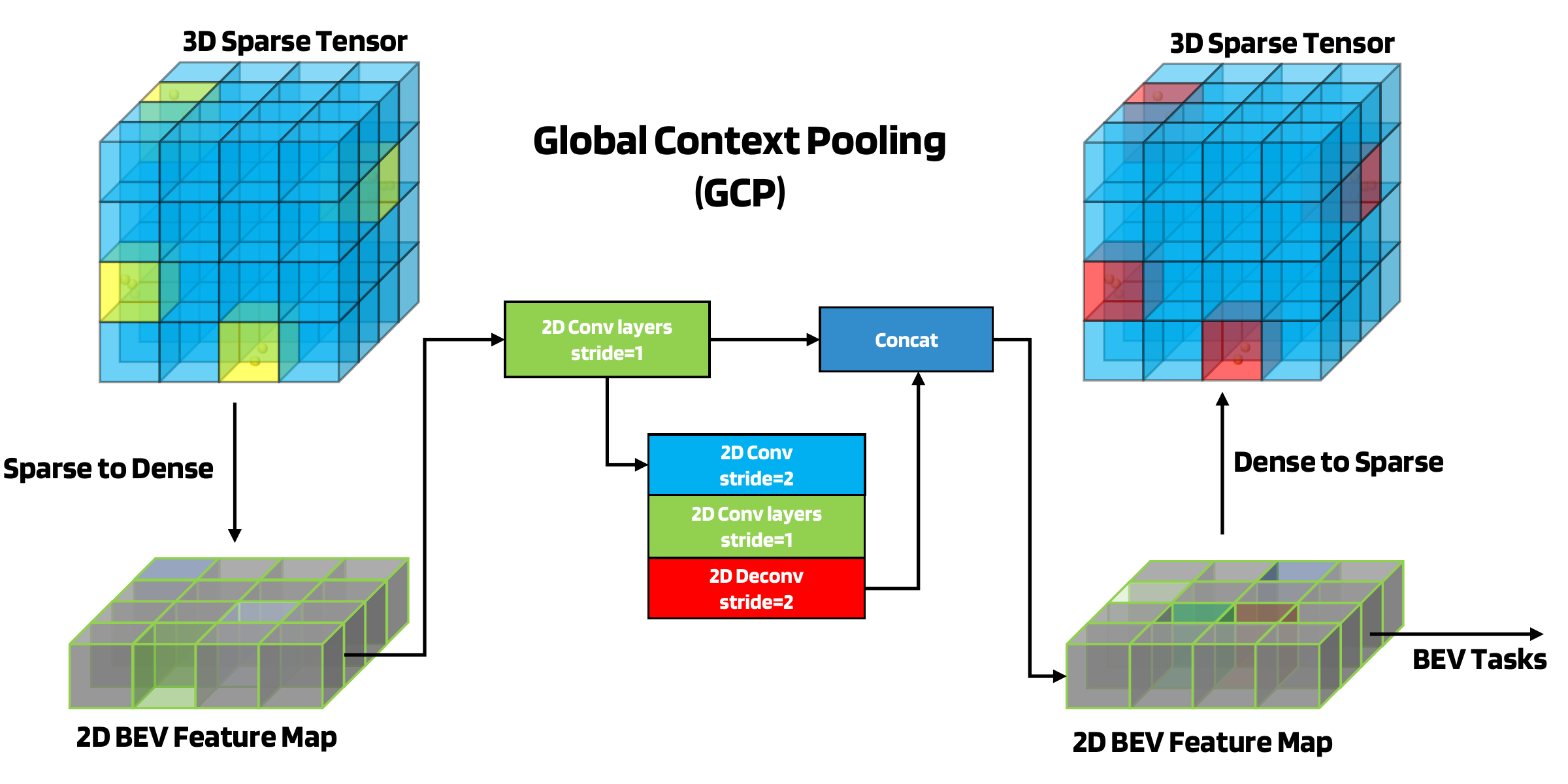}
  \caption{\textbf{Illustration of the Global Context Pooling (GCP) module}. 3D sparse tensor is projected to a 2D BEV feature map. Two levels of 2D BEV feature maps are concatenated and then converted back to a 3D sparse tensor, which serves as the input to the BEV task heads.}
  \label{fig:gcp}
\end{figure}

\begin{figure*}
\centering
\includegraphics[width=1.0\textwidth]{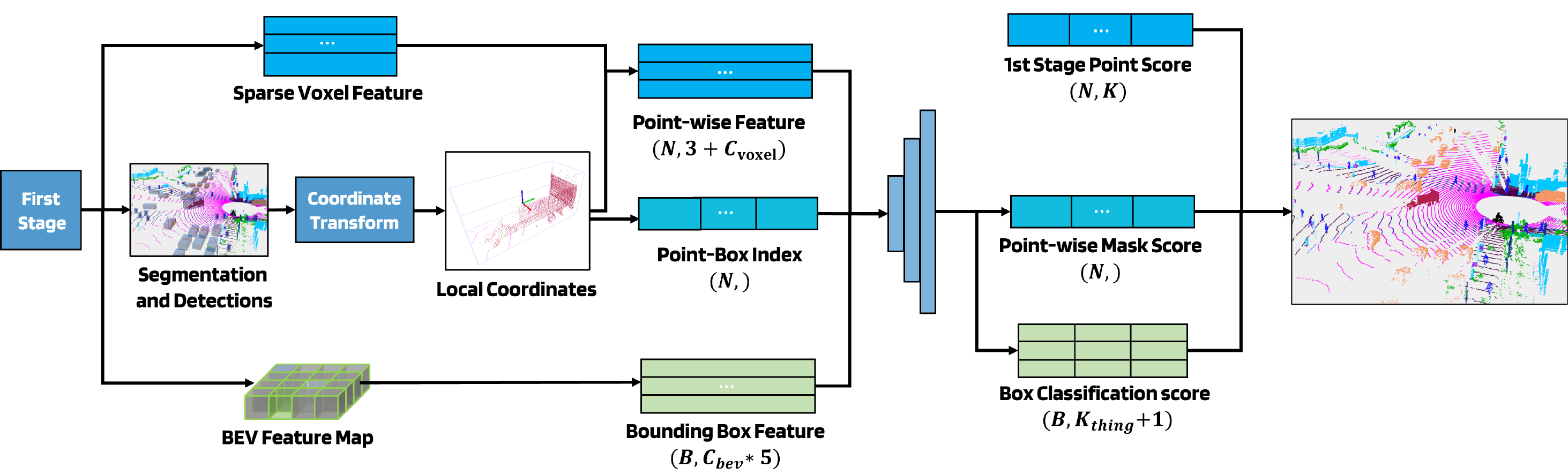}
\caption{\textbf{Illustration of the second-stage refinement pipeline.} The architecture of the second-stage refinement is point-based. We first fuse the detected boxes, voxel-wise features, and BEV features from the 1st stage to generate the inputs for the 2nd stage. The local coordinate transformation is applied to the points within each box. Then, a point-based backbone with MLPs, attention modules, and aggregation modules infer the box classification scores and point-wise mask scores. The final refined segmentation scores are computed by fusing the 1st and 2nd stage predictions.}
\label{fig:2nd_stage_refinement_illustration}
\end{figure*}

\subsection{Global Context Pooling}
3D sparse convolution drastically reduces the memory consumption of the 3D CNN for the LiDAR point cloud data, but it generally requires the layers of the same scale to retain the same sparsity in both encoder and decoder. This restricts the network to use only submanifold convolution~\cite{graham20183d} in the same scale. However, submanifold convolution cannot broadcast features to isolated voxels through stacking multiple convolution layers. This limits the ability of CNN to learn long-range global information. Inspired by the Region Proposal Network (RPN)~\cite{ren2015faster} in the 3D detection network, we design a Global Context Pooling (GCP) module to extract large-scale information through a dense BEV feature map. On the one hand, GCP can efficiently enlarge the receptive field of the network to learn global contextual information for the segmentation task. On the other hand, its 2D BEV dense feature can also be used for 3D object detection or other BEV tasks, by attaching task-specific heads with marginal additional computational cost.

As illustrated in Figure~\ref{fig:gcp}, given the low-resolution feature representation of the encoder output, we first transform the sparse voxel feature into a dense feature map $\mathcal{F}_{encoder}^{sparse}\in\mathbb{R}^{C'\times M'} \to \mathcal{F}^{dense}\in\mathbb{R}^{C'\times \frac{D}{d_{z}} \times \frac{H}{d_{x}} \times \frac{W}{d_{y}}}$, where $d$ is the downsampling ratio and $M'$ is the number of valid voxels in the last scale. We concatenate the features in different heights together to form a 2D BEV feature map $\mathcal{F}_{in}^{bev}\in\mathbb{R}^{(C'*\frac{D}{d_{z}}) \times \frac{H}{d_{x}} \times \frac{W}{d_{y}}}$. Then, we use a 2D multi-scale CNN to further extract long-range contextual information. Note that we can utilize a deeper and more complex structure with a trivial run-time overhead, since the BEV feature map has a relatively small resolution. Lastly, we reshape the encoded BEV feature representation to the dense voxel map, then transform it to the sparse voxel feature following the reverse dense to sparse conversion.

Benefiting from GCP, our architecture could significantly enlarge the receptive field, which plays an important role in semantic segmentation. In addition, the BEV feature maps in GCP can be shared with other tasks (eg. object detection) simply by attaching additional heads with slight increase of computational cost. By utilizing the BEV-level training like object detection, GCP can enhance the segmentation performance furthermore.

\subsection{Multi-task Training and Losses}
The 3D segmentation branch predicts voxel-level labels $L^{v}=\{l_{j}|l_{j}\in (1\dots K)\}_{j=1}^{M}$ given the learned voxel features $\mathcal{F}_{decoder}^{sparse}\in\mathbb{R}^{C\times M}$ output by the 3D decoder. $M$ stands for the number of active voxels in the output and $C$ represents the dimension of every output feature. We supervise it through a combination of cross-entropy loss and Lovasz loss~\cite{berman2018lovasz}: $\mathcal{L}_{SEG} = \mathcal{L}_{ce}^{v} + \mathcal{L}_{Lovasz}^{v}$. Note that $\mathcal{L}_{SEG}$ is a sparse loss, and the computational cost as well as the GPU memory usage are much smaller than dense loss.

The detection heads are applied on the 2D BEV feature map: $\mathcal{F}_{out}^{bev}\in\mathbb{R}^{C_{bev} \times \frac{H}{d_{x}} \times \frac{W}{d_{y}}}$. They predict a class-specific heatmap, the object dimensions and orientation, and a IoU rectification score, which are supervised by the focal loss~\cite{lin2017focal} ($\mathcal{L}_{hm}$) and L1 loss ($\mathcal{L}_{reg}, \mathcal{L}_{iou}$) respectively: $\mathcal{L}_{DET} = \lambda_{hm}\mathcal{L}_{hm}+\lambda_{reg}\mathcal{L}_{reg}+\lambda_{iou}\mathcal{L}_{iou}$, where the weights $\lambda_{hm}$, $\lambda_{reg}$, $\lambda_{iou}$ are empirically set to $[1,2,1]$.

During training, the BEV segmentation head is supervised with $\mathcal{L}_{BEV}$, a dense loss consisting of cross-entropy loss and Lovasz loss: $\mathcal{L}_{BEV} = \mathcal{L}_{ce}^{bev}+\mathcal{L}_{Lovasz}^{bev}$.

Our network is trained end-to-end for multiple tasks. Similar to \cite{lidarmtl2021}, we define the weight of each component of the final loss based on the uncertainty \cite{kendall2018multi} as follows: 
 \begin{equation}
\mathcal{L}_{total} =\sum_{{L}_{i}\in \left \{\substack{{L}_{ce}^{v},{L}_{Lovasz}^{v},{L}_{ce}^{bev},\\
{L}_{Lovasz}^{bev},{L}_{hm},{L}_{reg},{L}_{iou}}  \right \}
}\frac{1}{2\sigma^{2}_{i} }\mathcal{L}_{i}+\frac{1}{2} \log\sigma^{2}_{i} 
\end{equation}
where  $\mathcal\sigma_{i}$ is the learned parameter representing the degree of uncertainty in $task_{i}$. The more uncertain the $task_{i}$ is, the less $\mathcal{L}_{i}$ contributes to $\mathcal{L}_{total}$. The second part can be treated as a regularization term for $\mathcal\sigma_{i}$ during training.

Instead of assigning an uncertainty-based weight to every single loss, we first group the losses belonging to the same task with fixed weights. The resulting three task-specific losses (\textit{i.e.}, $\mathcal{L}_{SEG}$, $\mathcal{L}_{DET}$, $\mathcal{L}_{BEV}$) are then combined using weights defined based on the uncertainty: 
\begin{equation}
\mathcal{L}_{total} =\sum_{i\in \left \{SEG,DET,BEV  \right \}
}\frac{1}{2\sigma^{2}_{i} }\mathcal{L}_{i}+\frac{1}{2} \log\sigma^{2}_{i} 
\end{equation}

\subsection{Second-stage Refinement} 
\begin{figure*}[t]
\centering
\includegraphics[width=0.92\textwidth]{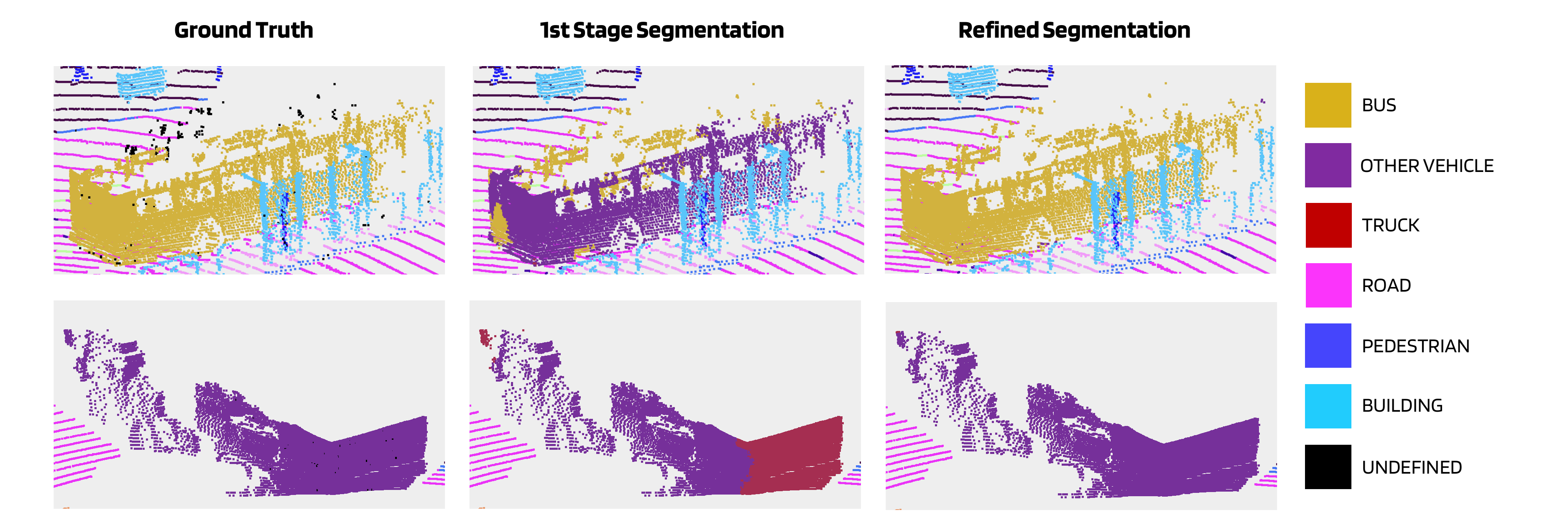}
\caption{\textbf{Examples of the 2nd-stage refinement}. The segmentation consistency of points of the thing objects can be improved by the 2nd stage.}
\label{fig:2nd_stage_refinement_example}
\end{figure*}

Coarse panoptic segmentation result can be obtained directly by fusing the first-stage semantic segmentation and object detection results,\textit{ i.e.}, assigning a unique ID to the points classified as one of the foreground \textit{thing} classes within a 3D bounding box. However, the points within a detected bounding box can be misclassified as multiple classes due to the lack of spatial prior knowledge, as shown in Figure~\ref{fig:2nd_stage_refinement_example}.  In order to improve the spatial consistency for the \textit{thing} classes, we propose a novel point-based approach as the second stage to refine the first-stage segmentation and provide accurate panoptic segmentation. 

The second stage is illustrated in Figure~\ref{fig:2nd_stage_refinement_illustration}. Specifically, it takes features from raw point cloud $P$, the $B$ predicted bounding boxes, sparse voxel features $\mathcal{F}_{decoder}^{sparse}$, and the BEV feature map $\mathcal{F}_{out}^{bev}$ to predict box classification scores $S_{box}$ and point-wise mask scores $S_{point}$. Given the $B$ bounding box predictions in the 1st stage, we first transform each point within a box into its local coordinates. Then we concatenate its local coordinates with the corresponding voxel features from $\mathcal{F}_{decoder}^{sparse}$.
Meanwhile, we extract 2nd-stage box-wise features as in \cite{yin2021center} from $\mathcal{F}^{bev}$. We assign a point-box index $I = \{{ind}_{i}|{ind}_{i}\in\mathbb{I}, 0 \leq {ind}_{i} \leq B\}_{i=1}^{N}$ to the points in each box. The points that are not in any boxes are assigned with index $\phi$ and will not be refined in the 2nd stage. Next, we use a PointNet-like network to predict point-wise mask scores ${S}_{point} = \{{sp}_{i}|{sp}_{i}\in (0, 1)\}_{i=1}^{N}$ and box classification scores ${S}_{box} = \{{sb}_{i}|{sb}_{i}\in {(0, 1)}^{{K}_{thing} + 1}\}_{i=1}^{B}$, where ${K}_{thing}$ denotes the number of \textit{thing} classes and the one additional class represents the remaining \textit{stuff} classes $\emptyset$. During training, we supervise the box-wise class scores through a cross-entropy loss and the point-wise mask scores through a binary cross-entropy loss.

We merge the 2nd-stage predictions with the 1st-stage semantic scores to generate the final semantic segmentation predictions $\hat{L}_{sem}$. To refine segmentation score $S_{2nd} = \{{rs}_{i}|{rs}_{i}\in {(0, 1)}^{{K}_{thing} + 1}\}_{i=1}^{N}$, we combine the point-wise mask scores with their corresponding box-wise class scores as follows:

\begin{equation}
S_{2nd}(j) = \begin{cases}
S_{point} \times S_{box}(j) ,\quad &{j} \in \mathcal{I}^{{K}_{thing}} \\
S_{point} \times S_{box}(j) + (1 - S_{point}) , \quad &{j} = \emptyset
\end{cases} 
\end{equation}
where ${K}_{thing}$ denotes the number of \textit{thing} classes, $\emptyset$ denotes the rest \textit{stuff} classes which would not be refined in the 2nd stage, ${S}_{point} = \{{sp}_{i}|{sp}_{i}\in (0, 1)\}_{i=1}^{N}$ is the point-wise mask scores, and ${S}_{box} = \{{sb}_{i}|{sb}_{i}\in {(0, 1)}^{{K}_{thing} + 1}\}_{i=1}^{B}$ is the box classification scores. $N$ and $B$ denote the number of points and boxes.

In addition, the points not in any boxes can be considered as $S_{2nd}(\emptyset) = 1$, which means their scores are the same as the 1st-stage scores. We then further combine the refined scores with the 1st-stage scores as follows:

\begin{equation}
S_{final} = \begin{cases}
S_{1st} \times S_{2nd}(\emptyset) + S_{2nd}(*) ,\quad &{ind}_{i} \ne \phi, * \ne \emptyset \\
S_{1st},\quad &{ind}_{i} = \phi
\end{cases} 
\end{equation}
where $\phi$ denotes the index where points are not in any boxes, and ${S}_{1st} = \{{sf}_{i}|{sf}_{i}\in {(0, 1)}^{K}\}_{i=1}^{N}$ is the 1st stage scores.

The scores $S_{final}$ are used to generate the semantic segmentation results $\hat{L}_{sem}$ through finding the class with the maximum score. It is intuitive to infer the final panoptic segmentation results through the 1st-stage boxes and the final semantic segmentation results $S_{box}$ and $\hat{L}_{sem}$. First, we extract points for a box where points and the box have the same semantic category. Then the extracted points will be assigned a unique index as the instance id for the panoptic segmentation.

\definecolor{Gray}{gray}{0.95}
\newcolumntype{g}{>{\columncolor{Gray}}c}

\section{Experiments}
In this section, we perform extensive tests of the proposed LidarMultiNet on five major benchmarks of the large-scale Waymo Open Dataset~\cite{sun2020scalability} (\textit{3D Object Detection }and \textit{3D Semantic Segmentation}) and nuScenes dataset~\cite{caesar2020nuscenes,fong2022panoptic} (\textit{Detection}, \textit{LiDAR Segmentation}, and \textit{Panoptic Segmentation}).

\subsection{Datasets and Metrics}

\noindent\textbf{Waymo Open Dataset (WOD)} contains 1150 sequences in total, split into 798 in the training set, 202 in the validation set, and 150 in the test set. Each sequence contains about 200 frames of LiDAR point cloud captured at 10 FPS with multiple LiDAR sensors. Object bounding box annotations are provided in each frame while the 3D semantic segmentation labels are provided only for sampled frames. WOD uses Average Precision Weighted by Heading (APH) as the main evaluation metric for the detection task. There are two levels of difficulty, LEVEL\_2 (L2) is assigned to examples where either the annotators label as hard or if the example has less than 5 LiDAR points, while LEVEL\_1 (L1) is assigned to the rest of the examples. Both L1 and L2 examples participate in the computation of the primary metric mAPH L2.  

For the semantic segmentation task, we use the v1.3.2 dataset, which contains 23,691 and 5,976 frames with semantic segmentation labels in the training set and validation set, respectively. There are a total of 2,982 frames in the final test set. WOD has semantic labels for a total of 23 classes, including an undefined class. Intersection Over Union (IOU) metric is used as the evaluation metric.\\

\noindent\textbf{NuScenes} contains 1000 scenes with 20 seconds duration each, split into 700 in the training set, 150 in the validation set, and 150 in the test set. The sensor suite contains a 32-beam LiDAR with 20Hz capture frequency. For the object detection task, the key samples are annotated at 2Hz with ground truth labels for 10 foreground object classes (\textit{thing}). For the semantic segmentation and panoptic segmentation tasks, every point in the keyframe is annotated using 6 more background classes (\textit{stuff}) in addition to the 10 \textit{thing} classes. NuScenes uses mean Average Precision (mAP) and NuScenes Detection Score (NDS) metrics for the detection task, mIoU and Panoptic Quality (PQ)~\cite{kirillov2019panoptic} metrics for the semantic and panoptic segmentation. Note that nuScenes panoptic task ignores the points that are included in more than one bounding box. As a result, the mIoU evaluation is typically higher than using full semantic labels.

\begin{table*}[t]
\centering
\resizebox{\linewidth}{!}{
\begin{threeparttable}
\begin{tabular}{l|>{\columncolor[gray]{0.95}}c|llllllllllllllllllllll}
\Xhline{4\arrayrulewidth}
\textbf{Waymo Leaderboard}                   & mIoU   & \rotatebox{80}{CAR}    & \rotatebox{80}{TRUCK}  & \rotatebox{80}{BUS}    & \rotatebox{80}{OTHERVEHICLE} & \rotatebox{80}{MOTORCYCLIST} & \rotatebox{80}{BICYCLIST} & \rotatebox{80}{PEDESTRIAN} & \rotatebox{80}{SIGN}   & \rotatebox{80}{TRAFFICLIGHT} & \rotatebox{80}{POLE}   & \rotatebox{80}{CONE} & \rotatebox{80}{BICYCLE} & \rotatebox{80}{MOTORCYCLE} & \rotatebox{80}{BUILDING} & \rotatebox{80}{VEGETATION} & \rotatebox{80}{TREETRUNK} & \rotatebox{80}{CURB}   & \rotatebox{80}{ROAD}   & \rotatebox{80}{LANEMARKER} & \rotatebox{80}{OTHERGROUND} & \rotatebox{80}{WALKABLE} & \rotatebox{80}{SIDEWALK} \\ \hline
\textbf{LidarMultiNet}             & \textbf{71.13} & \textbf{95.86} & \textbf{70.57} & 81.44 & 35.49         & 0.77       & \textbf{90.66}    & \textbf{93.23}     & \textbf{73.84} & \textbf{33.14}         & \textbf{81.37} & 64.68             & 69.92  & 76.73     & \textbf{97.37}   & \textbf{88.92}     & \textbf{73.58}      & \textbf{76.54} & 93.21 & \textbf{50.37}       & \textbf{53.82}        & \textbf{75.48}   & \textbf{87.95}   \\ 
\textbf{LidarMultiNet$^{\dagger}$}             & 69.69 & 95.47 & 65.36 & 80.44 & 31.60         & 1.23       & 90.04    & 92.70     & 71.82 & 31.54         & 79.57 & 61.39             & 69.30  & 76.78     & 97.17   & 88.47     & 71.49      & 74.64 & 92.72 & 48.34      & 51.92       & 74.51   & 86.75   \\ \Xhline{2\arrayrulewidth}
SegNet3DV2       & 70.48 & 95.73 & 69.03 & 79.74 & \textbf{37.0}           & 0            & 88.77    & 92.66     & 71.82 & 30.02         & 80.85 & \textbf{65.97}             & 69.53  & \textbf{76.97}     & 97.15   & 88.18     & 72.76      & 76.4  & \textbf{93.27} & 49.49       & 52.61        & 75.4    & 87.25   \\ 
HorizonSegExpert & 69.44 & 95.55 & 68.93 & \textbf{84.38} & 29.39         & 0.95       & 86.06    & 92.22     & 72.71 & 32.27         & 80.56 & 55.67             & 68.56  & 71.7      & 97.18   & 87.81     & 72.49      & 76.0   & 93.09 & 48.91       & 52.21        & 73.78   & 87.29   \\ 
HRI\_HZ\_SMRPV   & 69.38 & 95.79 & 66.97 & 78.36 & 32.8          & 0.04       & 89.06    & 91.83     & 72.93 & 31.73         & 79.82 & 61.33             & \textbf{70.23}  & 76.02     & 96.94   & 87.2      & 72.15      & 73.8  & 92.6  & 48.02       & 50.82        & 71.94   & 86.03   \\ 
Waymo\_3DSEG     & 68.99 & 95.47 & 69.54 & 77.85 & 31.29         & 0.67       & 84.4     & 91.12     & 72.73 & 31.45         & 80.12 & 61.25             & 68.87  & 74.87     & 96.96   & 87.89     & 72.93      & 74.62 & 92.61 & 44.68       & 49.33        & 73.1    & 86.13   \\ 
SPVCNN++         & 67.7  & 95.12 & 67.73 & 75.61 & 35.16         & 0            & 85.1     & 91.57     & 73.24 & 31.94         & 78.85 & 61.4              & 65.97  & 73.6      & 90.82   & 86.58     & 70.54      & 75.53 & 91.73 & 41.31       & 40.22        & 71.5    & 85.87   \\ 
PolarFuse        & 67.28 & 95.06 & 67.79 & 77.04 & 30.54         & 0            & 79.08    & 89.61     & 65.37 & 30.38         & 78.53 & 57.3              & 62.97  & 67.96     & 96.38   & 87.07     & 70.26      & 75.1  & 92.54 & 48.91       & 51.75        & 71.39   & 85.02   \\ 
LeapNet          & 66.89 & 94.45 & 65.68 & 79.01 & 30.74         & 0.01       & 79.44    & 90.07     & 70.46 & 30.34         & 77.67 & 52.71             & 61.58  & 66.84     & 96.65   & 87.01     & 69.8       & 74.49 & 92.61 & 44.91       & 47.17        & 73.59   & 86.34   \\ 
3DSEG                     & 66.77 & 94.64 & 66.95 & 77.61 & 30.67         & \textbf{2.11}       & 78.59    & 89.18     & 70.01 & 31.0           & 77.4  & 56.73             & 60.81  & 67.91     & 96.55   & 87.33     & 70.89      & 70.92 & 91.79 & 42.99       & 46.78        & 72.95   & 85.04   \\ 
CAVPercep                 & 63.73 & 93.64 & 62.83 & 68.12 & 22.83         & 1.89       & 71.73    & 87.39     & 67.03 & 29.41         & 74.29 & 55.03             & 55.6   & 60.84     & 96.15   & 86.48     & 68.06      & 68.33 & 91.24 & 41.26       & 46.19        & 70.55   & 83.15   \\ 
VS-Concord3D              & 63.54 & 92.6  & 66.9  & 73.13 & 24.26         & 0.83       & 76.77    & 85.51     & 66.88 & 29.98         & 75.81 & 53.86             & 62.26  & 68.11     & 86.04   & 75.13     & 66.95      & 67.95 & 90.6  & 40.43       & 45.43        & 65.6    & 82.96   \\ 
RGBV-RP Net               & 62.61 & 94.87 & 67.48 & 74.91 & 33.09         & 0            & 77.34    & 88.66     & 68.05 & 28.68         & 74.78 & 37.69             & 53.82  & 64.92     & 96.59   & 86.47     & 67.18      & 70.93 & 90.97 & 23.7        & 24.01        & 68.89   & 84.48  \\ 
\Xhline{4\arrayrulewidth}
\end{tabular}
\end{threeparttable}
}
\caption{Waymo Open Dataset Semantic Segmentation Leaderboard.
Our \textbf{LidarMultiNet} reached the highest mIoU of 71.13 and achieved the best accuracy on 15 out of the 22 classes. $\dagger$: without TTA and model ensemble.}
\label{table:leaderboard}
\end{table*}

\subsection{Implementation Details}
On the Waymo Open Dataset, the point cloud range is set to $[-75.2m, 75.2m]$ for $x$ axis and $y$ axis, and $[-2m, 4m]$ for the $z$ axis, and the voxel size is set to $(0.1m, 0.1m, 0.15m)$. Following \cite{yin2021center}, we transform the past two LiDAR frames using the vehicle's pose information and merge them with the current LiDAR frame to produce a denser point cloud and append a timestamp feature to each LiDAR point. Points of past LiDAR frames participate in the voxel feature computation but do not contribute to the loss calculation.

On the nuScenes dataset, the point cloud range is set to $[-54m, 54m]$ for $x$ axis and $y$ axis, and $[-5m, 3m]$ for the $z$ axis, and the voxel size is set to $(0.075m, 0.075m, 0.2m)$. Following the common practice~\cite{caesar2020nuscenes, yin2021center} on nuScenes, we transform and concatenate points from the past 9 frames with the current point cloud to generate a denser point cloud. Following \cite{yin2021center}, we apply separate detection heads in the detection branch for different categories.

During training, we employ data augmentation which includes standard random flipping, and global scaling, rotation and translation. We also adopt the ground-truth sampling~\cite{yan2018second} with the fade strategy~\cite{wang2021pointaugmenting}. We train the models using AdamW \cite{loshchilov2017decoupled} optimizer with one-cycle learning rate policy \cite{sylvain2018onecycle}, with a max learning rate of 3e-3, a weight decay of 0.01, and a momentum ranging from 0.85 to 0.95. We use a batch size of 2 on each of the 8 A100 GPUs. For the one-stage model, we train the models from scratch for 20 epochs. For the two-stage model, we freeze the 1st stage and finetune the 2nd stage for 6 epochs.

\begin{table}[t]
    \centering
    \resizebox{\linewidth}{!}{
    \begin{tabular}{ccccccccc}
    \Xhline{4\arrayrulewidth}
    Baseline & Multi-frame & GCP & $\mathcal{L}_{BEV}$ & $\mathcal{L}_{DET}$ & Two-Stage & TTA & Ensemble & mIoU                           \\\hline
    $\checkmark$ & & & & & & & & 69.90   \\
    $\checkmark$ & $\checkmark$ & & & & & & & 70.49 \\
    $\checkmark$ & $\checkmark$ & $\checkmark$ & & & & & &  71.43 \\
    $\checkmark$ & $\checkmark$ & $\checkmark$ &$\checkmark$ & & & & & 71.58 \\
    $\checkmark$ & $\checkmark$ & $\checkmark$ &$\checkmark$ &$\checkmark$ & & & & 72.06 \\
    $\checkmark$ & $\checkmark$ & $\checkmark$ &$\checkmark$ & $\checkmark$ &$\checkmark$ & & & 72.40 \\
    $\checkmark$ & $\checkmark$ & $\checkmark$ &$\checkmark$ & $\checkmark$ &$\checkmark$ & $\checkmark$ & & 73.05 \\
    $\checkmark$ & $\checkmark$ & $\checkmark$ &$\checkmark$ & $\checkmark$ &$\checkmark$ & $\checkmark$ & $\checkmark$ & 73.78 \\
    \Xhline{4\arrayrulewidth}
    \end{tabular}
    }
    \caption{Ablation studies for 3D semantic segmentation on the WOD \textit{validation} set. We show the improvement introduced by each component compared to our LidarMultiNet base network.}
    \label{table:ablation}
\end{table}

\subsection{Waymo Open Dataset Results} 
\subsubsection{3D Semantic Segmentation Challenge Leaderboard}

We tested the performance of \lidarmultinet on the WOD 3D Semantic Segmentation Challenge. Since the semantic segmentation challenge only considers semantic segmentation accuracy, our model is trained with focus on semantic segmentation, while object detection and BEV segmentation both serve as the auxiliary tasks. Since there is no runtime constraint, most participants employed the Test-Time Augmentation (TTA) and model ensemble to further improve the performance of their methods. For details regarding the TTA and ensemble, please refer to the supplementary material. Table~\ref{table:leaderboard} is the final WOD semantic segmentation leaderboard and shows that our \lidarmultinet~\cite{dongqiangzi2022lidarmultinet} achieves a mIoU of 71.13 and ranks the \nth{1} place on the leaderboard\footnote[1]{\url{https://waymo.com/open/challenges/2022/3d-semantic-segmentation}, accessed on August 06, 2022.}, and also has the best IoU for 15 out of the total 22 classes. Note that our \lidarmultinet uses only LiDAR point cloud as input, while some other entries on the leaderboard (e.g. SegNet3DV2) use both LiDAR points and camera images and therefore require running additional 2D CNNs to extract image features. 

For a better reference, we also test the result of \lidarmultinet that is trained with both detection and segmentation as the main tasks. \lidarmultinet reaches a mIoU of 69.69 on WOD 3D segmentation test set without TTA and model ensemble.

\subsubsection{Ablation Study on the 3D Semantic Segmentation Validation Set} 
We ablate each component of the LidarMultiNet and the results on the 3D semantic segmentation validation set are shown in Table~\ref{table:ablation}. Our baseline network reaches a mIoU of 69.90 on the validation set. On top of this baseline, multi-frame input (\textit{i.e.}, including the past two frames) brings a 0.59 mIoU improvement. The GCP further improves the mIoU by 0.94. The auxiliary losses, (\textit{i.e.}, BEV segmentation and 3D object detection) result in a total improvement of 0.63 mIoU, and the 2nd-stage improves the mIoU by 0.34, forming our best single model on the WOD validation set. TTA and ensemble further improve the mIoU to 73.05 and 73.78, respectively. 

\begin{table}[t]
    \centering
    \resizebox{\linewidth}{!}{
    \begin{tabular}{l|c|c|>{\columncolor[gray]{0.95}}c|ccc}
    \Xhline{4\arrayrulewidth}
        Methods & Modal & Frames & mAPH L2 & Veh. & Ped. & Cyc.\\ \hline
        PV-RCNN++~\cite{shi2021pv} & L& 1 & 70.20 & 73.47 & 69.00 & 68.15\\
        CenterPoint++ (Yin et al.~\citeyear{yin2021center})& L & 3 & 72.82 & 75.05 & 72.41 & 71.01\\
        SST\_3f~\cite{fan2022embracing}& L & 3 & 72.81 & 72.74 & 73.51 & 72.17\\
        AFDetV2~\cite{hu2022afdetv2}& L & 2 & 73.12  & 73.89 & 72.41 & 73.04 \\
        DeepFusion \cite{li2022deepfusion} & CL & 5 & 75.54 &  75.69 & 76.40 & 74.51 \\
        MPPNet~\cite{chen2022mppnet}& L & 16 & 75.67 & 76.91  & 75.93 & 74.18 \\
        CenterFormer~\cite{Zhou_centerformer}& L  & 16 & 76.29 & \textbf{78.28}  & \textbf{77.42} & 73.17 \\
        BEVFusion~\cite{liu2022bevfusion}& CL  & 3 & 76.33 & 77.48  & 76.41 & 75.09 \\ \hline
        LidarMultiNet (Ours)  & L &3 & \textbf{76.35} & 77.28 & 76.56 & \textbf{75.20}\\
    \Xhline{4\arrayrulewidth}
    \end{tabular}
    }
    \caption{Single-model detection performance comparisons on Waymo \textit{test} set. ``L" indicates LiDAR-only, and ``CL" denotes camera and LiDAR fusion. 
    }
    \label{table:waymo_detection_test}
\end{table}

\begin{table}[t]
    \centering
    \resizebox{\linewidth}{!}{
    \begin{tabular}{l|c|c|>{\columncolor[gray]{0.95}}c|ccc}
    \Xhline{4\arrayrulewidth}
        Methods & Modal & Frames & mAPH L2 & Veh. & Ped. & Cyc.\\ \hline
        PointPillar \cite{lang2019pointpillars}  & L & 1 & 57.53 & 62.50 & 50.20 & 59.90\\
        LiDAR R-CNN (Li et al.~\citeyear{li2021lidar})  & L & 1 & 60.10 & 64.20 & 51.70 & 64.40 \\
        PointAugmenting \cite{wang2021pointaugmenting} & CL & 1 & 66.7 & 62.70 &  70.55 & 74.41 \\
        CenterPoint (Yin et al.~\citeyear{yin2021center}) & L & 4 & 67.96 & 67.49 & 67.62 & 68.77 \\
        PV-RCNN++ \cite{shi2021pv} & L & 1 & 68.61 & 69.91 & 66.30 & 69.62 \\
        AFDetV2-Lite \cite{hu2022afdetv2} & L & 1 & 68.77 & 69.22 & 66.95 & 70.12 \\
        DeepFusion \cite{li2022deepfusion} & CL & 5 & - &  72.4 & 76.0 & - \\
        SST\_3f \cite{fan2022embracing} & L & 3 & 72.37 & 66.52 & \textbf{76.17} & 73.59  \\ \hline
        LidarMultiNet (Ours)  & L &3 & \textbf{75.15} & \textbf{73.38} & 74.10 & \textbf{77.96}\\
    \Xhline{4\arrayrulewidth}
    \end{tabular}
    }
    \caption{Detection performance comparisons on Waymo \textit{validation} set. ``L" indicates LiDAR-only, and ``CL" denotes camera and LiDAR fusion. 
    }
    \label{table:waymo_detection_val}
\end{table}

\begin{table}[t]
    \centering
    \resizebox{0.7\linewidth}{!}{
    \begin{tabular}{l|c|c}
    \Xhline{4\arrayrulewidth}
        Models & L2 mAPH & mIoU\\ \hline
        Detection-only & 75.01 & - \\
        Segmentation-only & - & 71.58 \\
        Joint Training & 75.15 & 71.93 \\
    \Xhline{4\arrayrulewidth}
    \end{tabular}
    }
    \caption{Comparison of the first-stage result of the jointly trained model and indepedently trained single-task models.}
    \label{table:ablation_study_jointly_training}
\end{table}

\begin{table}
    \begin{threeparttable}[t]
        \centering
        \resizebox{\linewidth}{!}{
        \begin{tabular}{l|l|c|>{\columncolor[gray]{0.95}}c|c>{\columncolor[gray]{0.95}}c|>{\columncolor[gray]{0.95}}cc}
        \Xhline{4\arrayrulewidth}
        ~ & Methods & Reference & mIoU & mAP & NDS & PQ & mIoU \\
        \hline
        \multirow{9}*{\rotatebox[origin=c]{90}{Segmentation}} & PolarNet & CVPR~\citeyear{Zhang_2020_CVPR} & 69.8 & - & - & - & - \\
        ~ & PolarStream & NeurIPS~\citeyear{chen2021polarstream} & 73.4 & - & - & - & - \\
        ~ & JS3C-Net & AAAI~\citeyear{yan2021sparse} & 73.4 & - & - & - & - \\
        ~ & AMVNet & IJCAIW~\citeyear{liong2020amvnet} & 77.3 & - & - & - & - \\
        ~ & SPVNAS & ECCV~\citeyear{tang2020searching} & 77.4 & - & - & - & - \\
        ~ & Cylinder3D++ & CVPR~\citeyear{zhu2021cylindrical} & 77.9 & - & - & - & - \\
        ~ & AF2S3Net & CVPR~\citeyear{Cheng_2021_CVPR} & 78.3 & - & - & - & - \\
        ~ & DRINet++ & arXiv~\citeyear{ye2021drinet++} & 80.4 & - & - & - & - \\
        ~ & SPVCNN++ & ECCV~\citeyear{tang2020searching} & 81.1 & - & - & - & - \\
        \hline
        \multirow{9}*{\rotatebox[origin=c]{90}{Detection}} & Cylinder3D & TPAMI~\citeyear{zhu2021cylinder3dtpami} & - & 50.6 & 61.6 & - & -\\
        ~ & CBGS & arXiv~\citeyear{zhu2019class} & - & 52.8 & 63.3 & - & -\\
        ~ & CenterPoint & CVPR~\citeyear{yin2021center} & - & 58.0 & 65.5 & - & -  \\
        ~ & HotSpotNet & ECCV~\citeyear{chen2020object} & - & 59.3 & 66.0 & - & - \\
        ~ & Object DGCNN & NeurIPS~\citeyear{wang2021object} & - & 58.7 & 66.1 & - & - \\
        ~ & AFDetV2 & AAAI~\citeyear{hu2022afdetv2} & - & 62.4 & 68.5 & - & - \\
        ~ & Focals Conv & CVPR~\citeyear{Chen2022focalsparse} & - & 63.8 & 70.0 & - & - \\
        ~ & TransFusion-L & CVPR~\citeyear{bai2022transfusion} & - & 65.5 & 70.2 & - & - \\
        ~ & LargeKernel3D & arXiv~\citeyear{chen2022scaling}  & - & 65.3 & 70.5 & - & - \\
        \hline
        \multirow{4}*{\rotatebox[origin=c]{90}{Panoptic}}  & EfficientLPS & T-RO~\citeyear{sirohi2021efficientlps} & - & - & - & 62.4 & 66.7\\
        ~ & Panoptic-PolarNet & CVPR~\citeyear{Zhou2021CVPR} & - & - & - & 63.6 & 67.0 \\
        ~ & SPVCNN++ & ECCV~\citeyear{tang2020searching} & - & - & - & 79.1 & 80.3 \\
        ~ & Panoptic-PHNet & CVPR~\citeyear{li2022panoptic} & - & - & - & 80.1 & 80.2 \\
        \hline
        ~ & LidarMultiNet & Ours & \textbf{81.4} & \textbf{67.0} & \textbf{71.6} & \textbf{81.4} & \textbf{84.0} \\
        \Xhline{4\arrayrulewidth}
        \end{tabular}
        }
    
        \caption{Comparison with state-of-the-art methods on the \textit{test} sets of three nuScenes benchmarks. A single \lidarmultinet model is used to generate predictions for all three tasks.}
        \label{table:nuscenes_test}
    \end{threeparttable}
\end{table}

\subsubsection{Evaluation on the 3D Object Detection Benchmark}
To demonstrate that \lidarmultinet can outperform single-task models on both detection and segmentation tasks, we tested it on the WOD 3D object detection benchmark and compared with state-of-the-art 3D object detection methods. The model is trained with both detection and segmentation as the main tasks, and its detection head is trained for detecting three classes, (\textit{i.e.} vehicle, pedestrian, and cyclist). The model is trained for 20 epochs with the fade strategy, \textit{i.e.}, ground-truth sampling for the object detection task is disabled for the last 5 epochs. The results on WOD test and validation set are shown in Table~\ref{table:waymo_detection_test} and Table~\ref{table:waymo_detection_val}. Our \lidarmultinet method reaches the highest mAPH L2 of 76.35 on the test set for a single model without TTA and outperforms the state-of-art 3D object detectors, including the multi-modal detectors which also leverage camera information. \lidarmultinet also outperforms other multi-frame fusion methods that require more past frames. Moreover, the same \lidarmultinet model reaches a mIoU of 71.93 on the WOD semantic segmentation validation set. In comparison, the other detectors and segmentation methods on the WOD benchmarks are all single-task models dedicated for either object detection or semantic segmentation.

\subsubsection{Effect of the Joint Multi-task Training}
Table~\ref{table:ablation_study_jointly_training} shows an ablation study comparing the first-stage result of the jointly-trained model with independently-trained models. The segmentation-only model removes the detection head, while keeping only the 3D segmentation head and the BEV segmentation head. The detection-only model keeps only the detection head. Compared to the single-task models, the jointly-trained model performs better on both segmentation and detection. In addition, by sharing part of the network among the tasks, the jointly-trained model is also more efficient than directly combining the independent single-task models.

\subsection{NuScenes Benchmarks}
\subsubsection{Comparison with State-of-the-art Methods}
On the nuScenes detection, semantic segmentation, and panoptic segmentation benchmarks, we compare \lidarmultinet with state-of-the-art LiDAR-based methods. The test set and validation set results of all three benchmarks are summarized in Table~\ref{table:nuscenes_test} and Table~\ref{table:nuscenes_val}, respectively. As shown in the tables, a single-model LidarMultiNet without TTA outperforms the previous state-of-the-art methods on each task. Combining the independently trained state-of-the-art single-task models (\textit{i.e.}, DRINet++, LargeKernel3D, and Panoptic-PHNet) reaches 80.4 mIOU, 70.5 NDS, and 80.2 PQ on the test set. In comparison, one single LidarMultiNet model without TTA outperforms their combined performance by 1.0\% in mIOU, 1.1\% in NDS, and 1.3\% in PQ.  

In summary, to the best of our knowledge, LidarMultiNet is the first time that a single LiDAR multi-task model surpasses previous single-task state-of-the-art methods for the three major LiDAR-based perception tasks.

\begin{table}
    \begin{threeparttable}[t]
        \centering
        \resizebox{\linewidth}{!}{
        \begin{tabular}{l|l|c|>{\columncolor[gray]{0.95}}c|c>{\columncolor[gray]{0.95}}c|>{\columncolor[gray]{0.95}}cc}
        \Xhline{4\arrayrulewidth}
        ~ & Methods & Reference & mIoU & mAP & NDS & PQ & mIoU \\
        \hline
        \multirow{5}*{\rotatebox[origin=c]{90}{Seg}} & PolarNet & CVPR~\citeyear{Zhang_2020_CVPR} & 71.0 & - & - & - & - \\
        ~ & SalsaNext & ISVC~\citeyear{cortinhal2020salsanext} & 72.2 & - & - & - & - \\
        ~ & Cylinder3D & CVPR~\citeyear{zhu2021cylindrical} & 76.1 & - & - & - & - \\
        ~ & AMVNet & IJCAIW~\citeyear{liong2020amvnet} & 77.2 & - & - & - & - \\
        ~ & RPVNet & ICCV~\citeyear{xu2021rpvnet} & 77.6 & - & - & - & - \\
        \hline
        \multirow{3}*{\rotatebox[origin=c]{90}{Det}} & CBGS & arXiv~\citeyear{zhu2019class} & - & 51.4 & 62.6 & - & -\\
        ~ & CenterPoint & CVPR~\citeyear{yin2021center} & - & 57.4 & 65.2 & - & -  \\
        ~ & TransFusion-L & CVPR~\citeyear{bai2022transfusion} & - & 60.0 & 66.8 & - & - \\
        \hline
        \multirow{3}*{\rotatebox[origin=c]{90}{Pan}}  & EfficientLPS & T-RO~\citeyear{sirohi2021efficientlps} & - & - & - & 62.0 & 65.6\\
        ~ & Panoptic-PolarNet & CVPR~\citeyear{Zhou2021CVPR} & - & - & - & 63.4 & 66.9 \\
        ~ & Panoptic-PHNet & CVPR~\citeyear{li2022panoptic} & - & - & - & 74.7 & 79.7 \\
        \hline
        ~ & LidarMultiNet & Ours & \textbf{82.0 }& \textbf{63.8} & \textbf{69.5} & \textbf{81.8} & \textbf{83.6} \\
        \Xhline{4\arrayrulewidth}
        \end{tabular}
        }
        
        \caption{Comparison with state-of-the-art methods on the \textit{validation} sets of three nuScenes benchamrks.}
        \label{table:nuscenes_val}
    \end{threeparttable}
\end{table}

\subsubsection{Effect of the Second-Stage Refinement}
An ablation study on the effect of the proposed 2nd stage is shown in Table~\ref{table:ablation_study_2nd_stage}. With the 1st-stage detection and semantic predictions, LidarMultiNet already can get high panoptic segmentation results by directly fusing these two results together. The proposed 2nd stage further improves both semantic segmentation and panoptic segmentation results.

\begin{table}[t]
    \centering
    \resizebox{0.8\linewidth}{!}{
    \begin{tabular}{lccccc}
    \Xhline{4\arrayrulewidth}
    \multirow{2}{*}{Stage} & Semantic  &  &  & Panoptic & \\ \cmidrule(rl){2-2} \cmidrule(rl){3-6}
         & mIoU & PQ & SQ & RQ & mIoU\\ \midrule
        1st stage & 81.7 & 81.2 & 90.5 & 89.3 & 83.3 \\
        2nd stage & 82.0 & 81.8 & 90.8 & 89.7 & 83.6 \\
    \Xhline{4\arrayrulewidth}
    \end{tabular}
    }
    \caption{Improvements of the 2nd-stage segmentation refinement on the nuScenes semantic segmentation and panoptic segmentation \textit{validation} sets.}
    \label{table:ablation_study_2nd_stage}
\end{table}

\subsubsection{Top-down LiDAR Panoptic Segmentation} CNN-based top-down panoptic segmentation methods have shown competitive performance compared to the bottom-up methods in the image domain. However, most previous LiDAR panoptic segmentation methods~\cite{Zhou2021CVPR,li2022panoptic} adopt the bottom-up design due to the need of an accurate semantic prediction. And the cumbersome network structures with multi-view or point-level features fusion make them difficult to perform well in the object detection task. On the other hand, thanks to the GCP module and joint training design, \lidarmultinet can reach top performance on both object detection and semantic segmentation tasks. Even without a 
dedicated panoptic head, \lidarmultinet already outperforms the previous state-of-the-art bottom-up method.

\section{Conclusion}
We present the \lidarmultinet, which reached the official \nth{1} place in the Waymo Open Dataset 3D semantic segmentation challenge 2022. \lidarmultinet is the first multi-task network to achieve state-of-the-art performance on all five major large-scale LiDAR perception benchmarks. We hope our LidarMultiNet can inspire future works in the unification of all LiDAR perception tasks in a single, versatile, and strong multi-task network. 

\appendix
\section{Appendix}

\subsection{Network Hyperparameters}
Hyperparameters of \lidarmultinet are summarized in  Table~\ref{table:design_space}.  Size of the input features of VFE is set to 16. The 3D encoder in our network has 4 stages of 3D sparse convolutions with increasing channel width 32, 64, 128, 256. The downsampling factor of the encoder is 8. The 3D decoder has 4 symmetrical stages of 3D sparse deconvolution blocks with decreasing channel widths 128, 64, 32, 32. 2D depth and width of the GCP module are set to 6, 6 and 128, 256, respectively.

\begin{table}[h]
    \centering
    \begin{tabular}{lcc}
    \Xhline{4\arrayrulewidth}
        Hyperparameters & Module & Values \\ \hline
        \#Input features & VFE & 16\\
        Voxel size & Voxelization & $(0.1m, 0.1m, 0.15m)$\\
        Range $x/y$ & Voxelization & $[-75.2m, 75.2m]$ \\
        Range $z$ & Voxelization & $[-2m, 4m]$ \\
        Downsampling & Encoder & 8 \\
        3D Depth & Encoder & 2, 3, 3, 3 \\ 
        3D Width & Encoder & 32, 64, 128, 256 \\ 
        3D Width & Decoder & 128, 64, 32, 32 \\ 
        2D Depth & GCP & 6, 6  \\
        2D Width & GCP &  128, 256 \\
        \Xhline{4\arrayrulewidth}
    \end{tabular}
    \caption{Implementation details of the LidarMultiNet}
    \label{table:design_space}
    \vspace{-2pt}
\end{table}

\subsection{Test-Time Augmentation and Ensemble}
In order to further improve the performance on the WOD semantic segmentation benchmark, we apply Test-Time Augmentation (TTA). Specifically, we use flipping with respect to $xz$-plane and $yz$-plane, [$0.95$, $1.05$] for global scaling, and [$\pm22.5\degree$, $\pm45\degree$, $\pm135\degree$, $\pm157.5\degree$, $180\degree$] for yaw rotation. Besides, we found pitch and roll rotation were helpful in the segmentation task, and we use $\pm8\degree$ for pitch rotation and $\pm5\degree$ for roll rotation. In addition, we also apply $\pm0.2m$ translation along $z$-axis for augmentation.

Besides the best single model, we also explored the network design space and designed multiple variants for model ensemble. For example, more models are trained with smaller voxel size $(0.05m, 0.05m, 0.15m)$, smaller downsample factor ($4\times$), different channel width (64), without the 2nd stage, or with more past frames (4 sweeps). For our submission to the leaderboard, a total of 7 models are ensembled to generate the segmentation result on the test set. 

\subsection{Runtime and Model Size} We tested the runtime and model size of LidarMultiNet on Nvidia A100 GPU. Compared with single-task models (nuScenes detection: 107ms, segmentation: 112ms, summation: 219ms), our multi-task network shows notable efficiency (nuScenes the 1st stage runtime: 126ms, model size: 135M, the 2nd stage runtime: 9ms, model size: 4M, WOD the 1st stage runtime: 145ms, model size: 131M, the 2nd stage runtime: 18ms, model size: 4M)

\begin{figure*}[ht]
\centering
\includegraphics[width=0.9\textwidth]{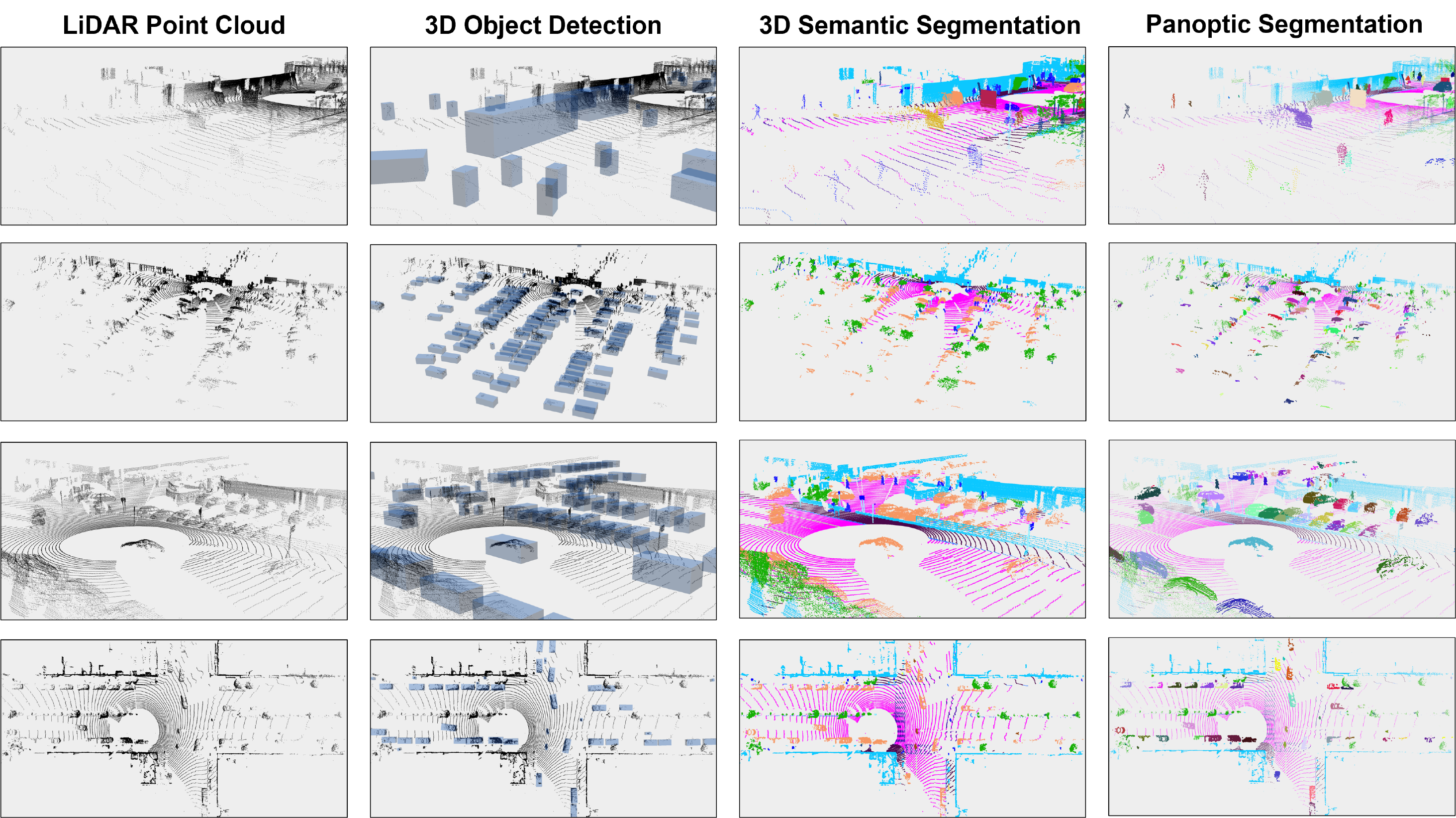}
\caption{\textbf{Example qualitative results of \lidarmultinet on the Waymo Open Dataset}.}
\label{fig:viz_waymo_val}
\end{figure*}

\begin{figure*}[ht]
\centering
\includegraphics[width=0.9\textwidth]{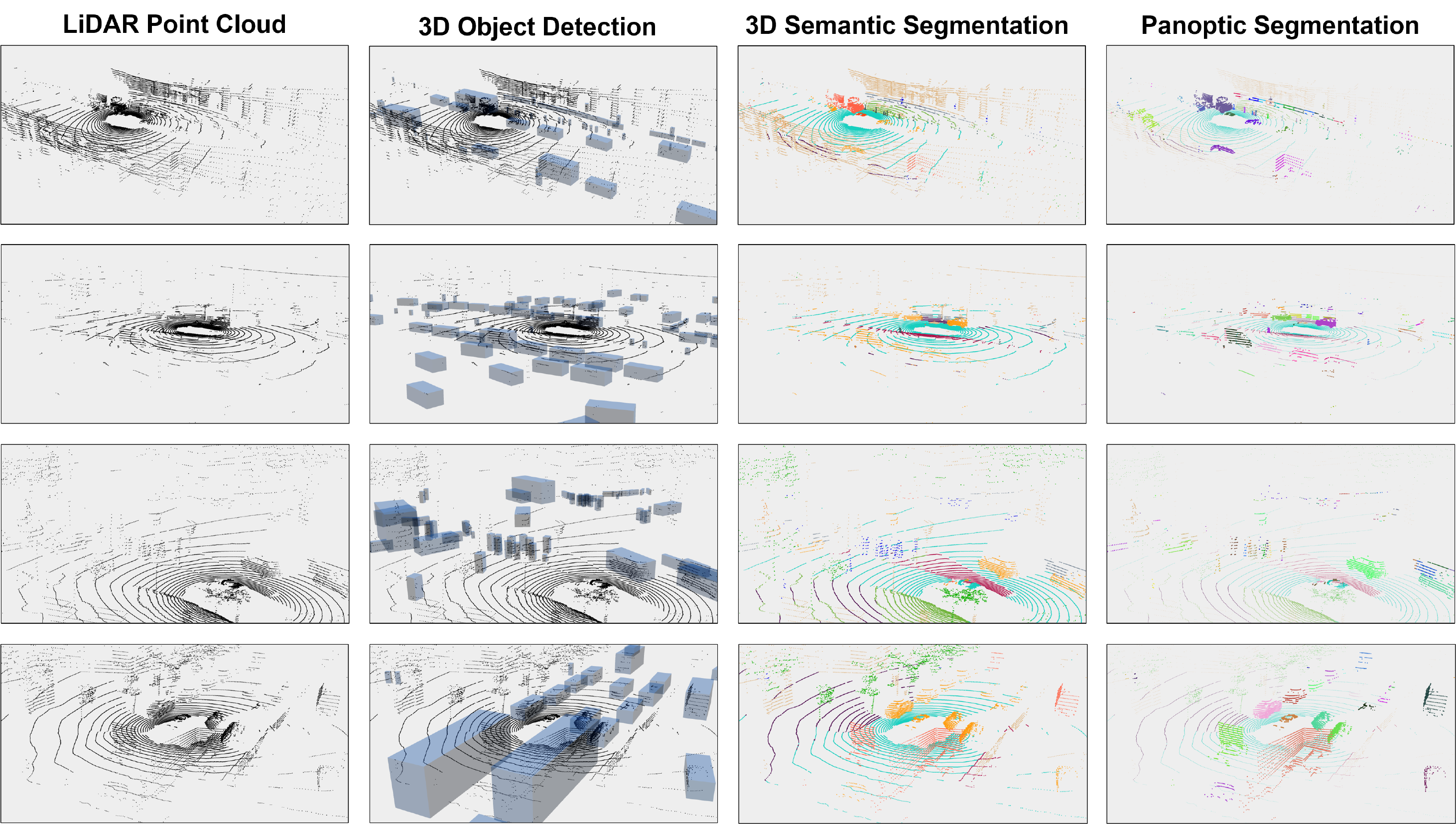}
\caption{\textbf{Example qualitative results of \lidarmultinet on the nuScenes dataset}.}
\label{fig:viz_nusc_val}
\end{figure*}

\subsection{Visualizations}
Some example qualitative results of \lidarmultinet on the validation sets of Waymo Open Dataset and nuScenes dataset are visualized in Figure~\ref{fig:viz_waymo_val} and Figure~\ref{fig:viz_nusc_val}, respectively. 

\bibliography{egbib}

\end{document}